\documentclass[10pt,twocolumn,letterpaper]{article}

\usepackage{cvpr}              

\usepackage{graphicx}
\usepackage{amsmath}
\usepackage{amssymb}
\usepackage{booktabs}

\usepackage{diagbox}
\usepackage{multirow}
\usepackage{array}

\usepackage[pagebackref,breaklinks,colorlinks]{hyperref}

\usepackage[capitalize]{cleveref}
\crefname{section}{Sec.}{Secs.}
\Crefname{section}{Section}{Sections}
\Crefname{table}{Table}{Tables}
\crefname{table}{Tab.}{Tabs.}

\def\cvprPaperID{4000} 
\def\confName{CVPR}
\def\confYear{2022}

\begin{document}

\title{Structure Information is the Key: Self-Attention RoI Feature Extractor in 3D Object Detection}

\author{Diankun Zhang\textsuperscript{1,2,3} , Zhijie Zheng\textsuperscript{1,2,3} , Xueting Bi\textsuperscript{4} , Xiaojun Liu\textsuperscript{1,2} \\
\textsuperscript{1} Key Laboratory of Electromagnetic Radiation and
Sensing Technology, Chinese Academy of Sciences\\
\textsuperscript{2} Aerospace information Research Institute, Chinese Academy of Sciences\\
\textsuperscript{3} University of Chinese Academy of Sciences\\
\textsuperscript{4} State Key Laboratory of Advanced Optical Communication Systems and Networks,\\ Department of Electronics, Peking University \\
{\tt\small \{zhangdiankun19,zhengzhijie19\}@mails.ucas.edu.cn, bixueting@stu.pku.edu.cn, lxjdr@mail.ie.ac.cn}
}
\maketitle

\begin{abstract}
   Unlike 2D object detection where all RoI features come from grid pixels, the RoI feature extraction of 3D point cloud object detection is more diverse. In this paper, we first compare and analyze the differences in structure and performance between the two state-of-the-art models PV-RCNN and Voxel-RCNN. Then, we find that the performance gap between the two models does not come from point information, but structural information. The voxel features contain more structural information because they do quantization instead of downsampling to point cloud so that they can contain basically the complete information of the whole point cloud. The stronger structural information in voxel features makes the detector have higher performance in our experiments even if the voxel features don't have accurate location information. Then, we propose that structural information is the key to 3D object detection. Based on the above conclusion, we propose a Self-Attention RoI Feature Extractor (SARFE) to enhance structural information of the feature extracted from 3D proposals. SARFE is a plug-and-play module that can be easily used on existing 3D detectors. Our SARFE is evaluated on both KITTI dataset and Waymo Open dataset. With the newly introduced SARFE, we improve the performance of the state-of-the-art 3D detectors by a large margin in \textit{cyclist} on KITTI dataset while keeping real-time capability. The code will be released at \href{https://github.com/Poley97/SARFE}{https://github.com/Poley97/SARFE}.
\end{abstract}

\section{Introduction}
\label{sec:intro}

3D perception is an important part of the automatic driving system \cite{grigorescu2020survey,bansal2018chauffeurnet}. 
At present, many 3D detectors have been proposed. The difference between point clouds and images makes the 2D object detection method \cite{ren2015faster,liu2016ssd,lin2017focal,cai2018cascade} not easy to directly extend to the point cloud data. To learn discriminative features from point-cloud, which is irregular and sparse, a common method is to convert the point cloud into a regular grid through voxelization, which will be efficiently processed by traditional Convolutional Neural Network (CNN). This is similar to the processing method that has been widely studied for 2D images, but the quantization loss of position information will inevitably occur in the voxelization process. These are called voxel-based approaches \cite{chen2017multi,zhou2018voxelnet,zheng2020cia,zheng2021se,deng2020voxel,lang2019pointpillars,yan2018second}. On the other hand, with the PointNet and PointNet++ proposed by Qi et al. \cite{qi2017pointnet,qi2017pointnet++}, some method directly learn features from raw point-cloud and predict 3D bounding boxes by foreground points and their features. These are called point-based methods \cite{qi2018frustum,shi2019pointrcnn,yang2019std,shi2020points,li2021lidar}. The point-based method can preserve accurate location information and have a flexible receptive field through some radius-based or knn-based aggregation operations instead of convolution. However, because the number of points in a point cloud is very large ($10k-200k$), to ensure the efficiency of detectors, the commonly used method is the farthest point sampling (FPS), which will also cause some information loss.

The voxel-based methods with predefined anchors have faster processing speed and can generate proposals with a higher recall rate, while the point-based methods can provide precise location information. In recent years, some methods have been proposed to fuse the advantages of voxel and point processing at the same time \cite{shi2020pv,shi2021pv,noh2021hvpr,bhattacharyya2020deformable}. PV-RCNN \cite{shi2020pv} is a representative work among them, and its performance has reached the state of the arts, but due to the introduction of complex point operations, it cannot meet the real-time requirements (10 fps). However, Voxel-RCNN \cite{deng2020voxel} proposed that the traditional voxel-based methods which extract features from bird’s eye view (BEV) ignore the 3D information, which reduces the performance of the detector. Therefore, the voxel RoI pooling operation is proposed, and the performance of the \textit{cat} is similar or even better than that of PV-RCNN on KITTI dataset \cite{geiger2013vision}. Meanwhile, it meets the real-time requirements (25 fps on an NVIDIA RTX 2080Ti GPU). 

PV-RCNN \cite{shi2020pv} and Voxel-RCNN \cite{deng2020voxel} both follow a standard two-stage object detection pipeline. The architectures of the backbone and region proposal network (RPN) are the same for these two detectors. There are two main differences between them: (1) PV-RCNN has a point branch to get the point features and an extra loss for the foreground point segmentation task, while Voxel-RCNN doesn’t have them. (2) The features used for RoI pooling are different. Voxel-RCNN pooling features from the 3D voxel backbone while PV-RCNN pooling features from weighted point features from the point branch.

In this paper, we first investigate the differences between two RoI pooling methods: pooling from point features and pooling from voxel features. Inspired by \cite{zhang2020bridging,zhang2018single}, we also eliminated other inconsistencies between PV-RCNN\cite{shi2020pv} and Voxel-RCNN\cite{deng2020voxel} for a fair comparison. From the experimental results, it can be concluded that the performance difference between the two is mainly due to the bottleneck effect formed by the down-sampling point in the process of extracting point features, which makes the features extracted by each grid point in the proposal more similar, resulting in structural information loss. This bottleneck effect and structural information loss caused by downsampling are more obvious on objects with complex structures (e.g. \textit{cyclist}), resulting in the disadvantage of PV-RCNN in detecting these objects. Through experiments, we propose that rich structure information is more important for 3D object detection than accurate position information.

Inspired by that, we propose a plug-and-play RoI feature extractor called Self-Attention RoI Feature Extractor (SARFE) which can be directly used on PV-RCNN and Voxel-RCNN. SARFE uses the self-attention mechanism to weigh different local features in the proposal, making the structural features more obvious and helping the network to identify and locate objects more precisely. Experiments on KITTI dataset support our analysis and conclusions, SARFE improves PV-RCNN and Voxel-RCNN performance on \textit{cyclist} by a large margin by extracting stronger structural information, which shows that 3D structure information is very important for the detection of objects with complex structures. 

The main contributions of this work can be summarized as (1) We indicate the essential difference between point-based and voxel-based RoI feature extractors is the tradeoff between location and structure information. Then we point out that compared to accurate position information, structural information is more important for 3D object detection; (2) We propose a plug-and-play module called self-attention RoI Feature Extractor (SARFE) to automatically adjust local features in a proposal to get features with stronger structure information; (3) Our proposed method SARFE achieves newly the-state-of-art performance on \textit{cyclist} of KITTI dataset\cite{geiger2013vision} while keeping real-time ability. SARFE also achieves state-of-the-art performance on the large-scale Waymo Open dataset \cite{sun2020scalability}. 

\section{Related Work}
\label{sec:related}
Nowadays, many 3D object detection algorithms have been proposed. Its pipeline is basically consistent with the 2D object detection algorithm, which can be categorized into two types: (1) single-stage detectors and (2) two-stage detectors.

At present, many excellent single-stage 3D detectors have been proposed \cite{zhou2018voxelnet,yan2018second,graham20183d,liu2015sparse,lang2019pointpillars,zheng2020cia,he2020structure,zheng2021se,tarvainen2017mean}. Because the single-stage detectors do not need to extract features from RoIs, it has a faster inference speed, but a loss in accuracy.

Two-stage object detectors use a second stage to refine the first-stage predictions with region-proposal-aligned features. PointRCNN \cite{shi2019pointrcnn} uses PointNet \cite{qi2017pointnet} to fuse and aggregate feature in a proposal for refinements. STD \cite{yang2019std} coverts point features to a compact representation in a region-proposal to improve the performance. PV-RCNN \cite{shi2020pv} utilizes a voxel-based network to generate 3D proposals while applying a point-based network to extract the features inside it. Voxel-RCNN \cite{deng2020voxel} has a similar structure to PV-RCNN but a different RoI feature extractor which results in the different performance of the detector. Analysis of these two different RoI feature extractors is a part of this paper. 

Representation learning on point clouds is also a crucial part of object detection. Point-cloud is sparse, unregular data, thus it can’t use convolution directly. PointNet \cite{qi2017pointnet} and PointNet++ \cite{qi2017pointnet++} learn point-wise feature from the raw points directly. DGCNN \cite{wang2019dynamic} uses a dynamic graph to learn point features. PAConv \cite{xu2021paconv} adopts a weight bank to construct a dynamic kernel for better handle point-cloud data. PCT \cite{guo2021pct} utilizes self-attention to get a stronger feature of point-cloud. Inspired by PCT, we propose a Self-Attention RoI Feature Extractor (SAREF) in this paper. The SARFE is a plug-and-play module used on two-stage 3d detectors and will be introduced in \cref{sec:sarfe}.

\begin{figure*}[ht]
	\centering
	\includegraphics[width=0.9\linewidth]{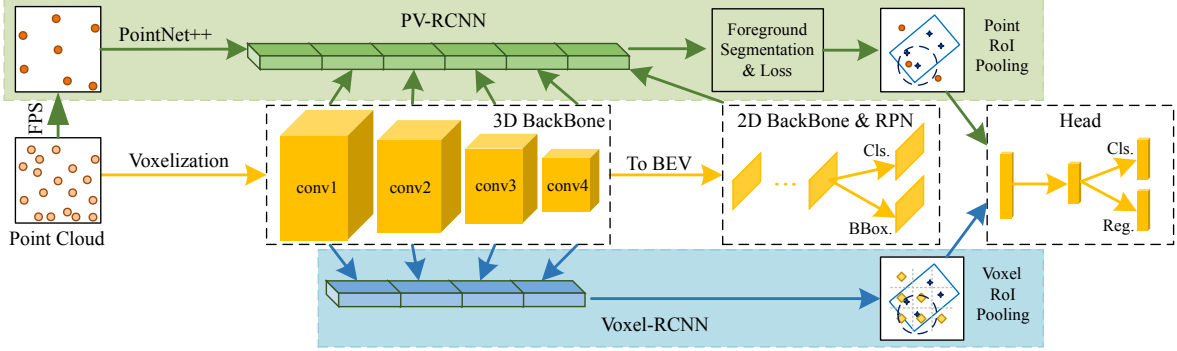}
	
	\caption{The difference between the architecture of PV-RCNN \cite{shi2020pv} and Voxel-RCNN \cite{deng2020voxel} . The main difference is the features used by RoI pooling: PV-RCNN uses point feature while Voxel-RCNN uses voxel features.}
	\label{fig:one}
\end{figure*}

\section{Essential Difference Between Voxel-based and Point-based RoI Feature Extractor}
\label{sec:analysis}

Due to the sparse, unorderly, and unregular characteristics of point cloud data, there are many different ways to process point cloud data. Previous work generally showed that point-based methods outperform voxel-based methods on object detection tasks. Voxel-RCNN proposed a viewpoint that the poor performance of the voxel-based method is due to the loss of 3D information in the process of extracting features in region proposals and proposed the voxel RoI pooling method. Therefore, Voxel-RCNN is used as a representative of the voxel-based RoI feature extractor method in this paper, because it can extract enough 3D information from the RoI just like the point-based method. PV-RCNN is one of the current SOTAs. It uses a point-based method to extract the features in the RoIs and has achieved excellent performance. This paper takes PV-RCNN as a representative of the point-based RoI feature extractor method. It is worth mentioning that the two models have the same architectures in the backbone and RPN, which makes the analysis fairer and more convenient.

\begin{table}[t]
	\small
	\centering
	\begin{tabular}{p{2.1cm}|p{0.55cm}p{0.55cm}p{0.55cm}|p{0.55cm}p{0.55cm}p{0.55cm}}
		\hline
		\multirow{2}{*}{Method} & \multicolumn{3}{c}{Car} &  \multicolumn{3}{c}{Cyclist} \\
		\cline{2-7}
		& Easy & Mod & Hard & Easy & Mod & Hard \\
		\hline
		PV-RCNN & \textbf{89.33}&	83.69&	\textbf{78.72}&		86.06&	69.47&	64.55 \\
		\hline
		\textbf{Voxel-RCNN*}: & & & & & & \\
		w/o PB & 89.29&	79.20&	78.40&		\textbf{89.58}&	\textbf{72.37}&	\textbf{68.19} \\
		w/ PB & 89.31 &	\textbf{84.34} &	78.72&	85.52&	72.01&	65.72\\
		\hline
	\end{tabular}
	\caption{The performance comparison between Voxel-RCNN and PV-RCNN after eliminating inconsistencies on the \textit{val} spilt of KITTI dataset \cite{geiger2013vision}. *: The detection classes of Voxel-RCNN are modified to 3 as same as the PV-RCNN. The results are reproduced by \cite{openpcdet2020}. PB = Point Branch of PV-RCNN.}
	\label{tab:one}
\end{table}

\subsection{Experiment Setting}
\label{sec:threeone}

All comparative experiments are conducted on the KITTI dataset \cite{geiger2013vision}. The dataset contains 7481 training samples and 7518 test samples. Following the common protocol, we divide the training samples into a \textit{train} split (3712 samples) and a \textit{val} split (3769 samples). We evaluate the results by average precision (AP) with different IoU threshold for different classes: 0.7 for \textit{car}, 0.5 for \textit{cyclist}. The difficulty levels in the evaluation: easy, moderate, and hard based on the object size, occlusion, and truncation levels. The models in this section are all trained in \textit{train} split, and the results obtained on the \textit{val} split.

The raw point clouds need to be transformed to regular voxels before being taken as input for both the backbone of Voxel-RCNN and PV-RCNN. As common settings, we clip the range of point-cloud into $[-3,1]m$ for $Z$ axis, $[-40,40]m$ for $Y$ axis, and $[0,70.4]m$ for $X$ axis. The input voxel size is set as $(0.05m,0.05m,0.10m)$ for both networks. 

In the training process, both networks are end-to-end optimized with Adam optimizer, 0.01 weight decay, and trained for 80 epochs on 8 RTX 3090 GPUs. We set the basic learning rate as 0.02 and batch size as 32. In terms of data augmentation, according to SECOND \cite{yan2018second}, we apply global random flipping, global scaling, global rotation, and GT database sampling for KITTI dataset.

During the inference process, we keep the top-128 proposals generated from 3D voxel CNN with a 3D IoU threshold of 0.7 for non-maximum-suppression(NMS). The proposals are further refined in the refinement stage with RoI features aggregated and augmented by SARFE. Finally, We use an NMS threshold of 0.1 to remove the redundant boxes.

\subsection{Difference elimination}
\label{sec:threetwo}

The frameworks of PV-RCNN and Voxel-RCNN are shown in \cref{fig:one}. Structurally, the two models are very similar. Both have the same architectures in the backbone and RPN. In the process of RoI feature extraction, the information of different scales in the backbone is also integrated. From \cref{fig:one}, it can be clearly seen that there are two main differences between the two models: (1) PV-RCNN has one point branch to extract the features of RoIs, which Voxel-RCNN doesn’t have. (2) PV-RCNN uses point features, while Voxel-RCNN uses voxel features to extract 3D features of RoIs.

It is worth noting that the Voxel-RCNN only give predictions for the \textit{car}, while the official source code of PV-RCNN can detect the \textit{car}, \textit{pedestrian} and \textit{cyclist} at the same time. For the fairness of comparison, we also modified the number of detection classes of Voxel-RCNN to 3 as the same as PV-RCNN. The modified Voxel-RCNN has the same head as PV-RCNN.

Voxel-RCNN does not need to use point features in the process of prediction, so it does not need the existence of point branches. However, it should be noted that there is an additional point segmentation loss on the point branch of PV-RCNN, which may affect the training. SA-SSD \cite{he2020structure} uses an auxiliary network that consists of a point branch to extract features from the voxel backbone, then performs point segmentation and center prediction, assists network training through auxiliary loss. Its structure is similar to the point branch of PV-RCNN, which shows that the point branch of PV-RCNN and corresponding segmentation loss may also play a role in assisting network training. Subsequent experiments also proved this point. To be fair, we use Voxel-RCNN with added point branching and segmentation loss to compare with PV-RCNN. The two networks are basically the same after the above adjustments, including the size of the voxel grid, the size of the anchor, the number of farthest point sampling (FPS) points, etc. The only difference between the two networks is that the source features used for the RoI feature extractor. The performances of the two models on the KITTI \textit{val} split are shown in \cref{tab:one}.

\subsection{Analysis}
\label{sec:threethree}

The previous series of excellent works \cite{yang2019std,yang20203dssd,shi2019pointrcnn,shi2020pv,he2020structure} make popular viewpoints generally consider that point features can provide more location information, and thus get more accurate object detection results. In theory, small objects have higher requirements for position information, because the same position error has a more severe impact on the IoU between the predicted bounding boxes and the ground truth of the small objects. However, it can be seen from \cref{tab:one} that compared with PV-RCNN, Voxel-RCNN obtains better performance on relatively small objects (\textit{cyclist}) and similar performance on large objects (\textit{car}) by extracting 3D voxel features instead of 3D point features in RoI feature extractor. This is contrary to the previous view that precise point information leads to performance degradation. When we add a point branch to the network for accurate location information supervision, the performance of Voxel-RCNN on \textit{car} is improved, while the performance on \textit{cyclist} is degraded.

Through experiments, we found that this difference in performance does not come from location information, but from structural information in the RoIs. This is also well understood intuitively that rich structural information is more conducive to the network's inference of the location and shape (size, ratio) of objects, especially for 3D objects with relatively fixed structures.

In the point cloud voxelization process, the voxel corresponding to each raw point must be in an activated (non-empty) state. This makes the number of activated voxels abundant (typically $>20k$ in both models). Through sparse convolution, the backbone will also convert the empty voxels around the active voxels into active, which is called submanifold dilation \cite{graham2017submanifold}, further increasing the number of activated voxels. In contrast, the number of sampling points obtained through FPS is fewer (2048 in PV-RCNN). Since FPS is approximate to uniform sampling in space, small objects like cyclists can only get fewer sampling points than large objects like cars, generally around a dozen. For objects with complex structures, like cyclists, too few sampling points make their structure information incomplete.

\begin{figure}[t]
	\centering
	\includegraphics[width=0.9\linewidth]{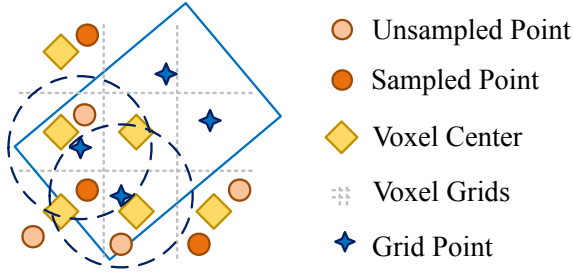}
	
	\caption{The essential difference between the two feature extraction methods is that the difference in the capability of aggregate structure information. There are only a few point features around grid points, which makes the local features aggregated by grid points less discriminative.}
	\label{fig:two}
\end{figure}

It should be noted that there are 6×6×6 grid points \cite{deng2020voxel,shi2020pv} in an RoI, which is far more than the number of sampling points. It is inevitable that multiple grid points have the same neighborhood sampling points, as shown in \cref{fig:two}. There are only the same FPS sampling points in the neighborhood of the two grid points, which means that the two grid points can only aggregate basically the same feature information. The smaller the feature difference between different grid points, the smaller the relationship between the extracted features in the proposal and their relative position in the proposal, which means the loss of structural information. On the contrary, due to the rich number of active voxels, the two grid points can aggregate different voxel features, making the features at different positions in RoI more discriminative, which means richer structure information.

In summary, there is a bottleneck effect in the method of aggregating local features from FPS sampling points. The bottleneck is the number of FPS sampling points, It can be alleviated by increasing the number of sampling points, but this will pay too much computational cost. This bottleneck makes the grid points unable to extract enough discriminative features, which limits the performance of the detector.

Through the above experiments, we have concluded that structural information is very important in 3D object detection. For a 3D object, local features (shape, size, location) are highly related. Intuitively, we can strengthen our cognition of objects in 3D space through the relative relationship (structural information) between multiple local features. Therefore, we proposed the SARFE in this paper. It strengthens the structure information extracted by the network to make the detector have better performance.

\section{Self-Attention RoI Feature Extractor}
\label{sec:sarfe}

\begin{figure*}[t]
	\centering
	\begin{subfigure}{0.72\linewidth}
		\includegraphics[height=5cm]{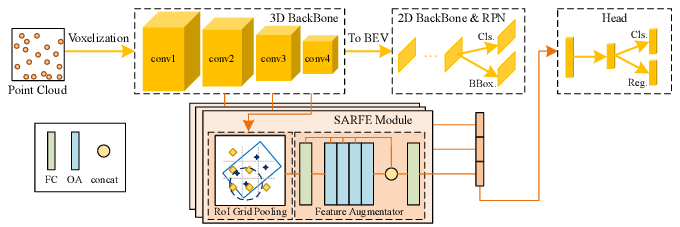}
		\caption{}
		\label{fig:three-a}
	\end{subfigure}
	\hfill
	\begin{subfigure}{0.12\linewidth}
		\includegraphics[height=5cm]{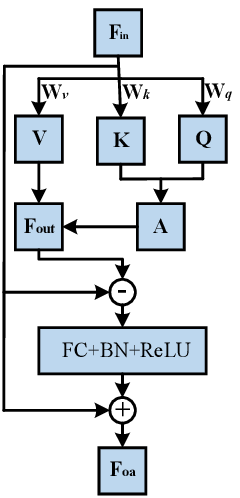}
		\caption{}
		\label{fig:three-b}
	\end{subfigure}
	\caption{(a) The overall architecture of SARFE with Voxel-RCNN framework. The SARFE module replaces the original RoI pooling module in Voxel-RCNN to get RoI features with stronger structure information; (b) The architecture of the offset-attention (OA) module proposed in \cite{guo2021pct}. In SARFE, we use this module to augment the structural information contained in RoI features through the relationship between local features.}
	\label{fig:three}
\end{figure*}
We propose a network for enhancing the structural information of features of each point in the region proposal called Self-Attention RoI Feature Extractor (SARFE), which is a plug-and-play network structure that can be used in two-stage 3D detectors such as Voxel-RCNN and PV-RCNN. 

The overall structure of SARFE is shown in \cref{fig:three-a}. SARFE mainly includes two parts, an RoI grid pooling module, and a self-attention local feature augmentator. Through the above experiments, we have concluded that structural information has a great influence on object detection (especially objects with complex structures like \textit{cyclist}). In daily life, we do not necessarily need to see the complete shape of an object, but only need to see some combination of local prominent structural features in order to obtain the class and location information. For example, through a bicycle wheel and handlebar, we can accurately infer the corresponding position of a bicycle. This shows that different local features of objects are related, and they can enhance or weaken each other. 

The self-attention \cite{vaswani2017attention} mechanism is very suitable for mining the associations between input features. However, unlike the application in natural language processing (NLP), the self-attention mechanism cannot be directly applied to point cloud features or voxel features. Because in NLP, each token has its basic semantics, while a single point doesn’t have. However, the local structure of the point cloud has basic semantic features. Therefore, we can regard the local features of the point cloud as a token in NLP and use self-attention to enhance it. We use an RoI grid pooling module to get the local features of grid points in RoIs, then a self-attention local feature augmentator is adopted to augment the structure information contained in these local features.

\textbf{RoI grid pooling}. This part is very similar to the RoI grid pooling methods used in Voxel-RCNN and PV-RCNN. The role of RoI grid pooling here is for the local features in the RoIs. The source of input features has many choices (such as voxel features or point features, depending on which model SARFE is used). SARFE can well augment both local features obtained by the aggregation of voxel features and point features, which will be described in \cref{sec:fivefour}. We uniformly sample 6×6×6 grid points in each 3D proposal, denoted as $\mathcal{G} = \left\{ {{g_i}\mid \;i = 1,...,216} \right\}$  , and then use the set abstraction operation \cite{qi2017pointnet++} to aggregate the surrounding points or voxel features to these grid points with multiple neighborhood radius as follows:

\begin{equation}
	{{\cal S}_k} = \left\{ {\left[ {{f_{nj}},{n_j} - {g_i}} \right]\;\;{\kern 1pt} \mid \;\;\;{\kern 1pt} ||{g_i} - {n_j}|| < {r_k}} \right\}
\end{equation}

where $n_j$  is the coordinate of the point or active voxel, $f_{nj}$  is the feature of the voxel or point corresponding to $n_j$ and $r_k$ is the neighborhood search radius. Then use an MLP and max pool to get the local feature $\mathcal{F}$  of grid points as follows:

\begin{equation}
	{\cal F} = \left\{ {{{\cal F}_k}\mid {{\cal F}_k} = {\rm{Maxpool}}({\rm{MLP(}}{{\cal S}_k}{\rm{)}})} \right\}
\end{equation}

RoI grid pooling can better capture the contextual information in the proposal, and at the same time have a more flexible receptive field. Compared with the 3D RoI-grid pooling operation in the other previous work \cite{shi2019pointrcnn,shi2019part}, this is more conducive for us to capture the local features of different parts of the object.

\textbf{Self-Attention local feature augmentator}. Through RoI gird pooling, each grid point contains the local features around it. We treat each grid point and its local features as a token. For each 3D proposal, we can get a sequence of 216 local features. Note that the self-attention mechanism is order-independent, so additional position encoding is needed in NLP, while point cloud itself is an order-independent data format, so no additional coding is required. 

We noticed that the self-attention mechanism has been successfully applied in the point cloud classification task \cite{guo2021pct}. Following that, we also adopt an offset-attention (OA) instead of the original self-attention mechanism in our local feature augmentator. The OA module is shown as \cref{fig:three-b}. For the input features  $\mathbf{F_{in}} \in {^{N \times d_i}}$ , Through the self-attention module with the same structure as in \cite{vaswani2017attention}, the output $\mathbf{F_{out}}$ can be obtained as follow:
\begin{equation}
	\mathbf{F_{out}}=\text{SelfAttention}(\mathbf{F_{in}})
\end{equation}
where  ${{\mathbf{F}}_{{\mathbf{out}}}}$ is the weighted feature of ${{\mathbf{F}}_{{\mathbf{in}}}}$. Finally, we can get the output of the OA module\cite{guo2021pct} ${{\mathbf{F}}_{{\mathbf{oa}}}}$  as follow:

\begin{equation}
	{{\mathbf{F}}_{{\mathbf{oa}}}} = {\text{ReLU}}({\text{BN}}({\text{Linear}}({\mathbf{F_{out}}} - {{\mathbf{F}}_{{\mathbf{in}}}}))) + {{\mathbf{F}}_{{\mathbf{in}}}}
\end{equation}

Each self-attention local feature augmentator is composed of $N$ offset-attention modules in series, the input of the first OA module is expressed as ${{\mathbf{F}}_{{\mathbf{oa,0}}}}$ and the output of each offset-Attention module is expressed as ${{\mathbf{F}}_{{\mathbf{oa}},{\mathbf{i}}}},i = 1,...,N$. We can get the final output of each self-attention local feature augmentator as
\begin{equation}
	{{\mathbf{F}}_{{\mathbf{local}}}} = \text{MLP}( {\text{Concat}}({{\mathbf{F}}_{{\mathbf{oa,0}}}},{{\mathbf{F}}_{{\mathbf{oa}},{\mathbf{1}}}},...,{{\mathbf{F}}_{{\mathbf{oa}},{\mathbf{N}}}}))
\end{equation}

In our experiments, we set $N=4$ for each feature source as shown in \cref{fig:three-a}.

~

\textbf{Multi-sources feature extraction}. Inspire by the multi-head mechanism proposed in \cite{vaswani2017attention}, we propose multi-sources independent feature extraction in our SARFE. Consider that there are multiple feature sources (e.g., multi-scale, voxel features and point features) input for the RoI feature extractor, the usual pipeline is to perform simple processing (e.g., interpolation and maxpooling) on the input features of different sources, concatenate them as one feature, then perform subsequent complex processing.
Because each feature source contains different feature information (for example, high-level features contain stronger semantic information, and low-level features contain more accurate location information). Concatenating features from different feature sources and then doing self-attention is not optimal, because it cannot ensure that the structural information of features from each source is enhanced at the same time. Therefore, we have prepared a SARFE module for each feature source to independently enhance the structural information they contain. Then the enhanced features will be concatenated as the feature of the RoIs. In our experiment, we use the features of \textit{conv2}, \textit{conv3}, and \textit{conv4} in backbone as three different feature sources as shown in  \cref{fig:three-a}.

\section{Experiments}
\label{sec:exp}
\begin{table*}
	\small
	\centering
	\begin{tabular}{l|l|ccc|ccc}
		\hline
		\multirow{2}{*}{Method} & \multirow{2}{*}{Reference} & \multicolumn{3}{c}{Car 3D Detection} & \multicolumn{3}{c}{Cyclist 3D Detection} \\
		\cline{3-8}
		& & Easy & Mod & Hard & Easy & Mod & Hard \\
		\hline
		SECOND \cite{yan2018second}	&Sensors 2019&	83.34&	72.55&	65.82&	71.33&	52.08&	45.83\\
		PointPillars \cite{lang2019pointpillars}&	CVPR 2019&	82.58&	74.31&	68.99&	77.10&	58.65&	51.92\\
		PointRCNN \cite{shi2019pointrcnn}&	CVPR 2019&	86.96&	75.64&	70.70&	74.96&	58.82&	52.53\\
		STD \cite{yang2019std} & CVPR2019 & 87.95 & 79.71& 75.09 & 78.69 & 61.59& 55.30 \\
		HotSpotNet \cite{chen2020object}&	ECCV2020&	87.60&	78.31&	73.34&	82.59&	65.95&	59.00\\
		Point-GNN \cite{shi2020point}&	CVPR2020&	88.33&	79.47&	72.29&	78.60&	63.48&	57.08\\
		3DSSD \cite{yang20203dssd}&	CVPR2020&	88.36&	79.57&	74.55&	82.45&	64.10&	56.90\\
		PV-RCNN \cite{shi2020pv}&	CVPR2020&	90.25&	81.43&	76.82&	78.60&	63.71&	57.65\\
		CIA-SSD \cite{zheng2020cia}&	AAAI 2021&	89.59&	80.28&	72.87&	-&	-&	-\\
		MGAF-3DSSD \cite{li2021anchor}&	MM’ 2021&	88.16 &79.68 &72.39&	80.64&	63.48&	57.08\\
		FromVoxelToPoint \cite{li2021voxel}&	MM’ 2021&88.53& 81.58& \textbf{77.37}&	81.49&	63.41&	56.40\\
		RangeIoUDet \cite{liang2021rangeioudet}& CVPR2021& 88.60 & 79.80 & 76.76 & 83.12 &67.77 & 60.26 \\
		Voxel-RCNN \cite{deng2020voxel}&	AAAI 2021&	\textbf{90.90}&	\textbf{81.62}&	{77.06}&	-&	-&	-\\
		\hline
		SARFE (ours)&	-&	88.88&	81.59	&76.74&	\textbf{84.88}&	\textbf{69.67} &	\textbf{62.26}\\
		\hline
	\end{tabular}
	\caption{Performance comparison with previous works on KITTI \textit{test} set. The results are evaluated by the mean Average Precision of 40 sampling recall points. Our SARFE shows great performance improvement in the \textit{cyclist} while maintaining high performance in the \textit{car}.}
	\label{tab:two}
\end{table*} 
In this section, we will introduce the implementation details of our SARFE framework in \cref{sec:fiveone} and show our experiment results on KITTI Dataset in \cref{sec:fivetwo} and results on Waymo Open Dataset in \cref{sec:fivethree}. Because our SARFE is a plug-and-play module rather than an end-to-end detection network, the proposed module is used in Voxel-RCNN framework to perform evaluations. To further validate our viewpoint and design, We analyze the evaluation results of using SARFE on PV-RCNN and Voxel-RCNN frameworks in \cref{sec:fivefour}.

\subsection{Experimental Setup}
\label{sec:fiveone}

\textbf{Network Architecture}. As shown in \cref{fig:three}. Our framework is based on Voxel-RCNN \cite{deng2020voxel}. The backbone and RPN of our network have the same structure as the official Voxel-RCNN while some hyperparameters are modified. Our 3D voxel CNN backbone has four levels with feature dimensions 16, 32, 64, 64. For each region proposal, as common setting in \cite{deng2020voxel,shi2020pv}, a  6×6×6 grid is used for RoI pooling. The neighboring radii for RoI pooling are 0.4m, 0.8m and 1.6m corresponding to feature sources \textit{conv2}, \textit{conv3} and \textit{conv4}. The number of the input and output channels of the offset-attention module is 32 as same as the number of channels of local features outputted from SARFE. Three SARFE modules are used to enhance the features from \textit{conv2}, \textit{conv3} and \textit{conv4} respectively, and then the local features from different sources will be concatenated for regression and classification.

\textbf{Datasets}. We evaluate our models on the well-known KITTI Dataset \cite{geiger2013vision} and newly Waymo Open Dataset \cite{sun2020scalability}. The settings of \textit{train} split and \textit{val} split are the same as setting in \cref{sec:threeone} on KITTI dataset. For the model submitted to the KITTI test server, we randomly selected 80\% of the training samples as the \textit{train} split and used the remaining 20\% as the \textit{val} split, which is also consistent with the usual settings. Waymo Open Dataset is a recently released and currently the largest dataset of 3D detection for autonomous driving. Different from KITTI that only provides annotations in camera FOV, the Waymo Open Dataset provides annotations for objects in the full $360^{\circ}$. There are 798 training sequences with around $158k$ LiDAR samples for training and 150 test sequences with around $30k$ LiDAR samples for evaluation.

\textbf{Training and Inference Detail}. The settings are the same as the settings in \cref{sec:threeone} while training on the KITTI Dataset. For Waymo Open Dataset, the detection range is $[-75.2,75.2]m$ for the $X$ and $Y$ axes and $[-2,4]m$ for the $Z$ axis, and we set the voxel size to $(0.1m,0.1m,0.15m)$.We train the entire
network with batch size 32, learning rate 0.003 for 36 epochs on 8 RTX 3090 GPUs. The other settings like data augmentation and NMS are as same as the settings on KITTI dataset.

\subsection{Results on KITII Dataset}
\label{sec:fivetwo}

\begin{table}
	\small
	\centering
	\begin{tabular}{l|p{0.66cm}p{0.66cm}p{0.66cm}p{0.66cm}p{0.66cm}p{0.66cm}}
		\hline
		\multirow{2}{*}{Method}	& \multicolumn{3}{c}{PV-RCNN\cite{shi2020pv}}  & \multicolumn{3}{c}{Voxel-RCNN*}\\
		& Easy & Mod & Hard & Easy & Mod & Hard\\
		\hline
		w/o SARFE & \textbf{86.06}&	69.47&	64.55 &89.58&	72.37&	68.19\\
		w/ SARFE  & 85.07&	\textbf{71.78}&	\textbf{67.18}  & \textbf{93.07}&	\textbf{73.93}	& \textbf{70.82}	\\
		\hline
		\textit{imporvement} & \textit{-0.99}&	\textit{+2.01}&	\textit{+2.63}  & \textit{+3.49}	& \textit{+1.56} & \textit{+2.63} \\
		\hline
	\end{tabular}
	\caption{Performance comparison with previous works on \textit{cyclist} class of KITTI \textit{val} split. The results are evaluated by the mean Average Precision of 11 sampling recall points. *: the results w/o SARFE are reproduced by \cite{openpcdet2020}.}
	\label{tab:three}
\end{table}

In order to verify the performance of the proposed model, we also submitted the results of our model on the \textit{test} set on the KITTI official test server. In order to further validate the effectiveness of our SARFE module, we also conducted evaluations on KITTI \textit{val} split. All result are evaluated by the mean average precision with a rotated IoU threshold 0.7 for \textit{car} and 0.5 for \textit{cyclist}. 

\textbf{Comparison with State-of-the-art Methods}. We compare the performance of other state-of-the-art by submitting and evaluating the detection results of our Voxel-RCNN+SARFE (donated as SARFE in \cref{tab:two}) framework on the KITTI test server. The results are shown in \cref{tab:two}. The SARFE network proposed in this paper has achieved excellent performance on \textit{cyclist} that improves the state-of-the-art performance by a large margin: increasing the mAP by 1.76\%, 1.90\%, 2.0\% on \textit{easy}, \textit{moderate}, and \textit{hard}. It should be noted that our model is trained for both \textit{car} and \textit{cyclist} detection, rather than separate models for each class.  For \textit{car}, our model can also achieve comparable performance with the state-of-the-art models, especially in moderate and hard levels. This shows that our model has a greater advantage in detecting difficult objects by introducing structural information. 

It can also be seen from \cref{tab:two} that our model has a very strong overall performance: while achieving new state-of-the-art performance in the \textit{cyclist}, it still has performance very close to the state-of-the-art in the \textit{car} (Voxel-RCNN\cite{deng2020voxel} is a detection model only for \textit{car}). None of the models in \cref{tab:two} can achieve close performance to our model on these two classes at the same time. We also report the performance on the KITTI \textit{val} split in \cref{tab:three}. It also shows that our model can greatly improve the state-of-the-art performance on \textit{cyclist} on multiple SOTA models.

Meanwhile, SARFE only brings a small additional computational cost, so that the model can maintain the original real-time performance. This part will be analyzed in \cref{sec:fivefour}.
\subsection{Results on the Waymo Open Dataset}
\label{sec:fivethree}

We also evaluate the performance of SARFE on the Waymo Open Dataset. The mean average precision (mAP) and the mean average precision weighted by heading (mAPH) are used for evaluation on Waymo Open Dataset. The test data are split into two difficulty levels, where Level 1 denotes the ground-truth objects with at least 5 inside points while Level 2 denotes the ground-truth objects with at least 1 inside point.

\textbf{Comparison with State-of-the-art Methods}. The performance of our SARFE and other state-of-the-art methods are shown in \cref{tab:four}. All method in \cref{tab:four} is performed on a single frame. Similar to the performance on KITTI dataset, we have maintained the performance on \textit{vehicle} and greatly improved the performance on \textit{cyclist}. 
\subsection{Results Analysis}
\label{sec:fivefour}

\textbf{Location Information vs. Structure Information}. In previous works, a common viewpoint is that using point features containing accurate location information for proposal refinement \cite{shi2019pointrcnn,shi2020pv} or supervising voxel feature learning \cite{he2020structure} can achieve better detection results. However, through experiments, we found that precise location information can indeed bring some help, but rich structural information reflects a more important role: it can greatly improve the performance of the network on objects with complex structures and can replace the role of precise location information. In \cref{tab:one}, the precise location information from the point branch improves the performance of Voxel-RCNN* to the same level of PV-RCNN on \textit{car} while degrading the performance of Voxel-RCNN* on \textit{cyclist}. From \cref{tab:two} we can see that with stronger structure information, our SARFE achieves the same, even better performance on \textit{car}, while a far better performance on \textit{cyclist} compare to PV-RCNN with precise location information.

This shows that structural information is more important than accurate location information in 3D object detection tasks. Unlike images, the objects in the 3D point cloud must contain structural information. The structural information composed of several local features is conducive to the network to identify and locate objects.

\textbf{Plug-and-Play module}. To further verify the effect of our proposed SARFE module and structure information on 3D detection, we also conducted experiments on PV-RCNN. Since PV-RCNN extracts the features of the proposal directly from the point features, we regard it as one feature source and use the neighborhood radius of 0.8 and 1.6 to aggregate the local information. The results are shown in \cref{tab:three}. Due to the lack of structural information of PV-RCNN (analyzed in \cref{sec:analysis}), SARFE improves its performance by a large margin on \textit{pedestrian} and \textit{cyclist}. It shows that the enhancement of structural information can greatly improve the performance of the 3D detectors, and it also shows that our SARFE is a plug-and-play module that can improve the detection performance of multiple models.

\textbf{Real-time capability}. As an object detection algorithm for autonomous driving applications, real-time capability is very important. Experiments show that our algorithm does not bring too much additional computational cost that can still maintain the real-time nature of the original model. As shown in \cref{tab:four}. With the addition of the SARFE module, the model has an acceptable reduction in FPS while a significant performance improvement. Finally, the entire model can still maintain performance far exceeding real-time requirements (32fps vs. 10fps).

\begin{table}
	\small
	\centering
	\begin{tabular}{ll|cccc}
		\hline
		\multirow{2}{*}{Difficulty} & \multirow{2}{*}{Method} & \multicolumn{2}{c}{Vehicle 3D} & \multicolumn{2}{c}{Cyclist 3D}\\
		& & mAP & mAPH & mAP & mAPH\\
		\hline
		\multirow{4}{*}{Level 1} &SECOND\cite{yan2018second} & 49.98& 49.47& 42.79& 42.36\\
		& PPBA \cite{ngiam2019starnet}& 67.52 & 67.00 & -& -\\
		& SA-SSD \cite{he2020structure}& 70.24	&69.54&	57.06&	55.45\\
		& RangeDet\cite{fan2021rangedet} & 75.83 & 75.83 & 64.59 & 63.08\\
		& SARFE & \textbf{78.19} & \textbf{77.77} & \textbf{70.28} & \textbf{68.97}\\
		\hline
		\multirow{4}{*}{Level 2} &SECOND\cite{yan2018second} & 42.79& 42.36& 35.33& 34.12 \\
		& PPBA \cite{ngiam2019starnet}& 59.55 & 59.09& -& -\\
		& SA-SSD \cite{he2020structure}& 	61.79&	61.17&	54.80&	53.25\\
		& RangeDet\cite{fan2021rangedet} & 67.12 & 66.73 & 61.93 & 60.49 \\
		& SARFE & \textbf{70.02} & \textbf{69.64} & \textbf{67.62} & \textbf{66.36} \\
		\hline

		\hline
	\end{tabular}
	\caption{Performance comparison with previous works on Waymo Open Dataset \textit{test} set for the \textit{vehicle} and \textit{cyclist}  detection.}
	\label{tab:four}
\end{table}

\begin{table}
	\small
	\centering
	\begin{tabular}{l|cc}
		\hline
		\multirow{2}{*}{Method} & \multicolumn{2}{c}{Frames per second}\\
		& w/o SARFE & w/ SARFE\\
		\hline
		PV-RCNN & 8.3 & 6.7\\
		Voxel-RCNN & 42.1 & 32.1\\
		\hline
	\end{tabular}
	\caption{The real-time capability of our SARFE module. The results are measured on KITTI Dataset by an RTX 3090 GPU. }
	\label{tab:five}
\end{table}

\section{Conclusion}
\label{sec:conclusion}

In this paper, we first analyze the difference in the structure and performance of the two SOTA 3D detection methods PV-RCNN and Voxel-RCNN, then propose that the reason for the performance gap between these two models is the difference in the ability of the network to capture structural information. Although the voxel feature can cause the location information to be blurred, it retains the rich structural information of the original data, while the point feature is just the opposite. Due to FPS, a large amount of structural information is lost, but accurate location information is retained. Our analysis and experiments show voxel features that contain richer structural information can achieve better detection performance. Then, we propose Self-Attention RoI Feature Extractor (SARFE) to further improve the network's ability to use structural information for proposal refinement. Our SARFE is a plug-and-play module that can be used on multiple existing models and greatly improve their detection performance for objects with a complex structure like cyclists. Our SARFE is evaluated on both KITTI dataset and Waymo Open dataset. SARFE achieves new state-of-arts performance on \textit{cyclist} while maintaining the high detection performance of \textit{car} on KITTI dataset and real-time capability.

\newpage
{\small
	\bibliographystyle{ieee_fullname}
	\bibliography{egbib}
}

\end{document}